\documentclass[conference]{IEEEtran}
\IEEEoverridecommandlockouts
\usepackage{cite}
\usepackage{amsmath,amssymb,amsfonts}
\usepackage{graphicx}
\usepackage{textcomp}
\usepackage{xcolor}
\usepackage{booktabs}
\usepackage{multirow}
\usepackage{float}
\usepackage{placeins}
\usepackage{capt-of}
\def\BibTeX{{\rm B\kern-.05em{\sc i\kern-.025em b}\kern-.08em
    T\kern-.1667em\lower.7ex\hbox{E}\kern-.125emX}}
\begin{document}

\title{Social Intuition vs. Machine Reasoning: Anticipating Human-Robot Interaction from multiple modalities}

\author{

\IEEEauthorblockN{Raphael Lorenzo-Louis}
\IEEEauthorblockA{
{Inria, CNRS, UL, Loria, HUCEBOT}\\
{F-54600 Villers-les-Nancy France}\\
{CEA, List, F-91120, Palaiseau, France}
}
\and
\IEEEauthorblockN{Bertrand Luvison}
\IEEEauthorblockA{
{Universit\'{e} Paris-Saclay} \\
{CEA, List, F-91120, Palaiseau, France}
}
\and
\IEEEauthorblockN{Serena Ivaldi}
\IEEEauthorblockA{
{Inria, CNRS, UL, Loria, HUCEBOT} \\
{F-54600 Villers-les-Nancy France}
}

}

\maketitle

\begin{abstract}

Anticipating whether a person will interact from one's own perspective is a highly intuitive task for humans, that relies
on a combination of cues. 

% Our work compares 2D-pose and video-based modalities for this task.
% We first assess human performance, then report results from fully trained 
% lightweight pose-based models and state-of-the-art vision-language models.

We investigate how humans perform at predicting a person's intention to interact from a service robot's point of view, using pose-only or full video
input, then benchmark different lightweight pose-based models and state-of-the-art vision-language models.

%, comparing binary interaction anticipation
% on a fixed pilot subset of 100 test tracks (25 positive, 75 negative) with a 1 second anticipation cut-off.

We conducted our benchmark on the \emph{HUI360} dataset on a fixed pilot subset of 100 test tracks (25 positive, 75 negative).
We found that with pose-only input, human annotators outperform lightweight trained pose models but not by large margins ($+0.08$ in F1-Score). 
But when given full egocentric video with a target bounding box, human annotators perform substantially better 
and largely outperform the Vision-Language Models ($+0.2$ in F1-Score). We also compared VLMs of different size and under different input conditions, and
found that the best results do not correlate with model size. Our result confirms that 
predicting interactions is a challenging task for social robots and that reasoning-capable models are necessary but their actual reasoning capabilities alone do not suffice
to match the social intuition of humans.

\end{abstract}

\begin{IEEEkeywords}
human-robot interaction, anticipation, egocentric vision, vision-language models, nonverbal cues
\end{IEEEkeywords}

\section{Introduction}
Social robots operating among people need reliable predictors of 
interaction and intention with enough anticipation to adequately plan motion and behaviors. 
In particular the \textit{binary} prediction of whether someone will or will not interact with
the robot is crucial for robots to address only those who are willing to interact.

% Egocentric 360$^\circ$ sensing captures rich social context, yet deployed pipelines frequently reduce video to lightweight 2D pose streams that can run on embedded hardware, while recent vision-language models (VLMs) promise flexible video reasoning at substantially higher compute and latency.

From a computer-vision perspective, recent progress in video understanding 
using vision-language models makes them natural candidates for this task, 
but they remain computationally expensive to run and suffer from high latency, whereas on the other hand 
the very lightweight 2D-pose modality can be processed
in real-time on edge devices.

We investigate how the interactant representation choice affects the prediction performance on the anticipation task, 
both for humans and vision models, and for the latter
how different classes of machine learning approaches perform, under different modalities.

We build on the \textbf{HUI360} dataset and pose-learning baselines~\cite{lorenzo2026hui360} 
with a fixed pilot benchmark: 100 test tracks (25 interaction-positive, 75 negative) 
and a one-second cut-off before the predicted interaction window.
Human annotators label each track under modalities: $\mathbf{Pose}$ (2D-pose overlay on a static scene context) and $\mathbf{V_{bbox}}$ 
(egocentric video with a target bounding box).

Baselines trained with the $\mathbf{Pose}$ modality include reference sequence models 
trained end-to-end on sequences: an LSTM baseline, ST-GCN~\cite{yan2018stgcn}, and SkateFormer~\cite{do2025skateformer}.

On the VLMs side we compare six models spanning on-device and API inference:
Qwen3.5-0.8B and InternVL3-1B (served locally),
Qwen3.5-35B-A3B and Qwen3.8-Max,
and Gemini~3.5-Flash-Lite and Gemini~3.6-Flash,
evaluated under three input conditions: $\mathbf{I}_c$ (image crop), $\mathbf{V}+\mathbf{I}_c$ (video with image reference), and $\mathbf{V_{bbox}}$ (video with per-frame bounding box).

We further ablate thinking mode, sampling rate, prompt length, and equirectangular \emph{vs.}\ pinhole-camera style projection on two representative models.

\section{Related Work}
\textbf{Interaction anticipation and datasets.}
HUI360~\cite{lorenzo2026hui360} is the largest in-the-wild dataset for egocentric human--robot interaction anticipation;
Earlier intent-prediction benchmarks are smaller or less reproducible.
The People Approaching Robots Database (PAR-D)~\cite{thompson2024pard} 
aggregates data from two robots in public environments, including \emph{Shutter}, a stationary robot photographer tracking human subjects
with Azure Kinect sensors and from the ATC dataset with a mobile service robot in a shopping mall \cite{katoMayIHelpYou}. 
Human performance was assessed on PAR-D from short 3D skeleton visualizations (0.2 to 4 seconds)
shown to crowd annotators (balanced samples). The results of \cite{thompson2024pard} establish average human F1-scores
between $0.65$ and $0.86$ with a noticeable variance of human accuracy depending on the observation length and the deployment context in each split.

\textbf{VLMs for social and anticipatory perception.}
Recent HRI studies use video VLMs as commonsense proxies 
for when to engage, often triggered by lightweight preamble detectors~\cite{bu2025vlmproxy}.
In social navigation, VLM-Social-Nav~\cite{song2024vlmsocialnavsociallyawarerobot} scores candidate robot actions with a vision-language model to produce socially compliant trajectories at inference time, without task-specific training on navigation datasets.
At the same time some works on affect perception with VLMs have shown that models still 
struggle with emotion recognition from videos, suffering from inherent and prompt-induced 
biases \cite{bhattacharyya2025evaluatingvisionlanguagemodelsemotion, agarwal2026visionlanguagemodelsstruggle} as well as 
the lack of high frequency temporal modelling \cite{agarwal2026visionlanguagemodelsstruggle}.
In \cite{parreira2026badidea} VLMs to human anticipatory judgment are compared on outcome prediction from 
contextual video and from observer's facial reactions, 
finding strong context-based performance but weaker use of social cues.

\section{Methods}
\subsection{Dataset}
\label{sec:dataset}

HUI360~\cite{lorenzo2026hui360} is an egocentric 360$^\circ$ dataset from Shelfy, a static tray-holding robot in nine indoor environments.
% The dataset further extends its data corpus with data from the SSUP dataset \cite{ssuphri} collected from a trashcan robot in public spaces.
Persons are detected and tracked with SAM2~\cite{ravi2024sam2segmentimages} on equirectangular frames; interactions are auto-annotated from masks and the tray zone, then manually reviewed.
The HUI split has 4310 samples (375 interactions), we do not use the complementary \textit{SSUP-A} split ($>$28{,}000 samples; a different robot and domain).
Body keypoints come from ViTPose~\cite{xu2022vitpose} (17) and Sapiens~\cite{khirodkar2024sapiens} (308); $\mathbf{Pose}$ models are retrained on HUI-Train with 17 COCO joints.

A \emph{track} is one followed person over a short window in the 360$^\circ$ view.
We evaluate 100 HUI-Test tracks (two environments unseen by the $\mathbf{Pose}$ models): 25 \emph{positive} (the person will physically pick or place an item on the tray) and 75 \emph{negative} (no such action).
Gaze, hesitations, and speech are cues, not labels.
Each window ends 16 frames ($\approx$1\,s at 15\,fps) before an interaction onset (pick or place something on the tray) for positives, largest apparent size (closest approach) for negatives : 
so negatives are harder samples rather than just people walking away.
% Figure~\ref{fig:sampling} illustrates the sampling strategy.

% \begin{figure}[t]
%   \centering
%   \fbox{\parbox[c][1.4in][c]{0.9\linewidth}{\centering
%   \textit{Placeholder: positive/negative track sampling relative to interaction onset and closest approach.}}}
%   \caption{Track sampling for interaction anticipation on HUI360-Test.
%   Positive windows end 16 frames before physical interaction; negative windows end 16 frames before the subject's largest apparent size.
%   \textcolor{blue}{TODO: add illustration}.}
%   \label{fig:sampling}
% \end{figure}

\subsection{Human baseline}

Six annotators judged the same 100 tracks under two conditions: \textbf{(1)} $\mathbf{Pose}$, a pose sequence overlaid on a static RGB scene, and \textbf{(2)} $\mathbf{V_{bbox}}$, 
egocentric video with a bounding box on the target person.
They were not required to complete both: three labeled both, two only $\mathbf{V_{bbox}}$, and one only $\mathbf{Pose}$ (hence $n{=}4$ and $n{=}5$ in Table~\ref{tab:preliminary}).
Each completed condition covers all 100 tracks, we do not omit trials that were hard to judge.

For each clip they answered: \textit{Will the person highlighted in the video take or place something 
on the robot's tray (camera wearer) within the next 2 seconds?}
Clips were shown in randomized order.
% Users could choose to speed-up the video up to twice the original speed. The average answering time takes into account
% the time spent watching the video and answering the question.
The static RGB context in $\mathbf{Pose}$ is required because the image is an equirectangular 360$^\circ$ projection: it grounds the skeleton in the scene (a corridor or hallway).

\begin{figure}[t]
  \centering
  \includegraphics[width=\linewidth]{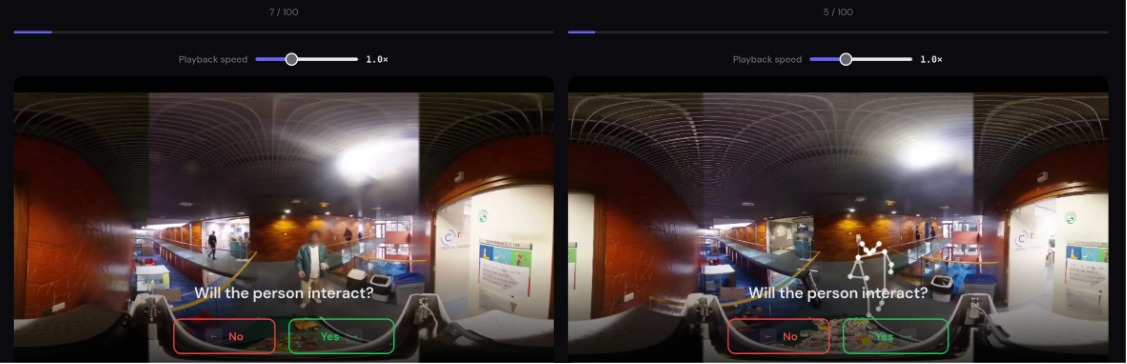}
  \caption{Human annotation interface for the two evaluation conditions: $\mathbf{V_{bbox}}$ (left, egocentric video) and $\mathbf{Pose}$ (right, skeleton overlay on static scene context). Annotators answer whether the highlighted person will interact within the next 2\,s.}
  \label{fig:human-setup}
\end{figure}

\subsection{Baseline skeleton models}

Table~\ref{tab:preliminary} reports scores for models trained on HUI360-Train with $\mathbf{Pose}$ input
and evaluated on the 100-track subset: (1) LSTM, (2) ST-GCN, and (3) SkateFormer.
All models take a fixed 32-frame sequence (at 15 fps) of 17 COCO body joints with three channels per joint (2D position and confidence).
(1) The LSTM baseline is a three-layer recurrent network (hidden size 128), resulting in a parameter count of just 0.37M.
(2) ST-GCN~\cite{yan2018stgcn} is a spatial-temporal graph convolutional network operating on the OpenPose skeleton graph with learnable edge-importance weighting, 
we used the default implementation from the official repository and convert our skeletons to the OpenPose format, ST-GCN has 3.07M trainable parameters.
(3) SkateFormer~\cite{do2025skateformer} is a skeletal-temporal transformer that applies self-attention over explicit spatial partitions of the skeleton with 
a parameter count of 1.91M.

Training follows the open-source HUI360 protocol with a 16-frame anticipation cutoff ($T_{\mathrm{adv}}$), weighted binary cross-entropy loss, 
AdamW optimizer (learning rate of $10^{-3}$, batch size is 128).
LSTM and ST-GCN train for 100 epochs (weight decay $5\times10^{-5}$ and $0.1$, respectively); 
SkateFormer trains for 600 epochs with weight decay $0.1$, and uses gradient clipping.
ST-GCN and SkateFormer use image-normalized joint coordinates without global standardization; 
the LSTM additionally normalizes joints within each 
bounding box and takes globally normalized bounding box coordinates as input.

Table \ref{tab:preliminary} reports the F1-score, accuracy, precision and recall (at $0.5$ probability threshold),
as well as per-sequence latency on an RTX~2000 Ada GPU for a full 32-frame sequence, excluding 2D-pose extraction, 
which we assume feasible in real time at 15\,fps with off-the-shelf estimators such as ViTPose~\cite{xu2022vitpose} 
or YOLO26-Pose~\cite{jocher2026ultralyticsyolo26unifiedrealtime}.

\subsection{Vision-language models}
\label{sec:methods-vlm}

We evaluate 
the open-weights Qwen3.5-0.8B, InternVL3-1B, Qwen3.5-35B-A3B, Qwen3.8-Max models as well as the closed ones Gemini~3.5-Flash-Lite, and Gemini~3.6-Flash
under three input conditions 
(denoted $\mathbf{I}_c$, $\mathbf{V}+\mathbf{I}_c$, and $\mathbf{V_{bbox}}$):
\begin{enumerate}
  \item $\mathbf{I}_c$: a cropped image of the target person at the last visible frame (16 frames before onset). Since we want a single image input,
  we assume the last image before the cutoff is the most meaningful for the model to estimate if someone will interact.
  \item $\mathbf{V}+\mathbf{I}_c$: the full equirectangular video and a reference image of the target person; 
  the model is asked to reason about the person shown in the image. Empirically, all models we tested succeeded at tracking the right 
  target person across frames, making this an adequate way of disambiguating multi-person scenarios. Additionally, as
  motivated for the $\mathbf{I}_c$ modality, this close-up image can be used to identify more subtle cues such as gaze and facial
  expression.
  \item $\mathbf{V_{bbox}}$: the full equirectangular video together with normalized (0-999) bounding box coordinates of the target person in the last frame. With this modality
  we exploit the "visual prompting" capacities of recent VLMs to refer to an object by passing normalized coordinates.
\end{enumerate}
Unless noted otherwise, all VLM runs use 8~FPS video sampling from 960x480 equirectangular videos but image crops are made 
directly on high resolution 3840x1920 equirectangular videos. All models are used with thinking mode enabled and a thinking budget of 1024 tokens.
API models are queried via DashScope (Qwen) and Google Gemini, on-device models are served with vLLM on an RTX~2000~Ada GPU.

\begin{figure}[t]
  \centering
  \includegraphics[width=\linewidth]{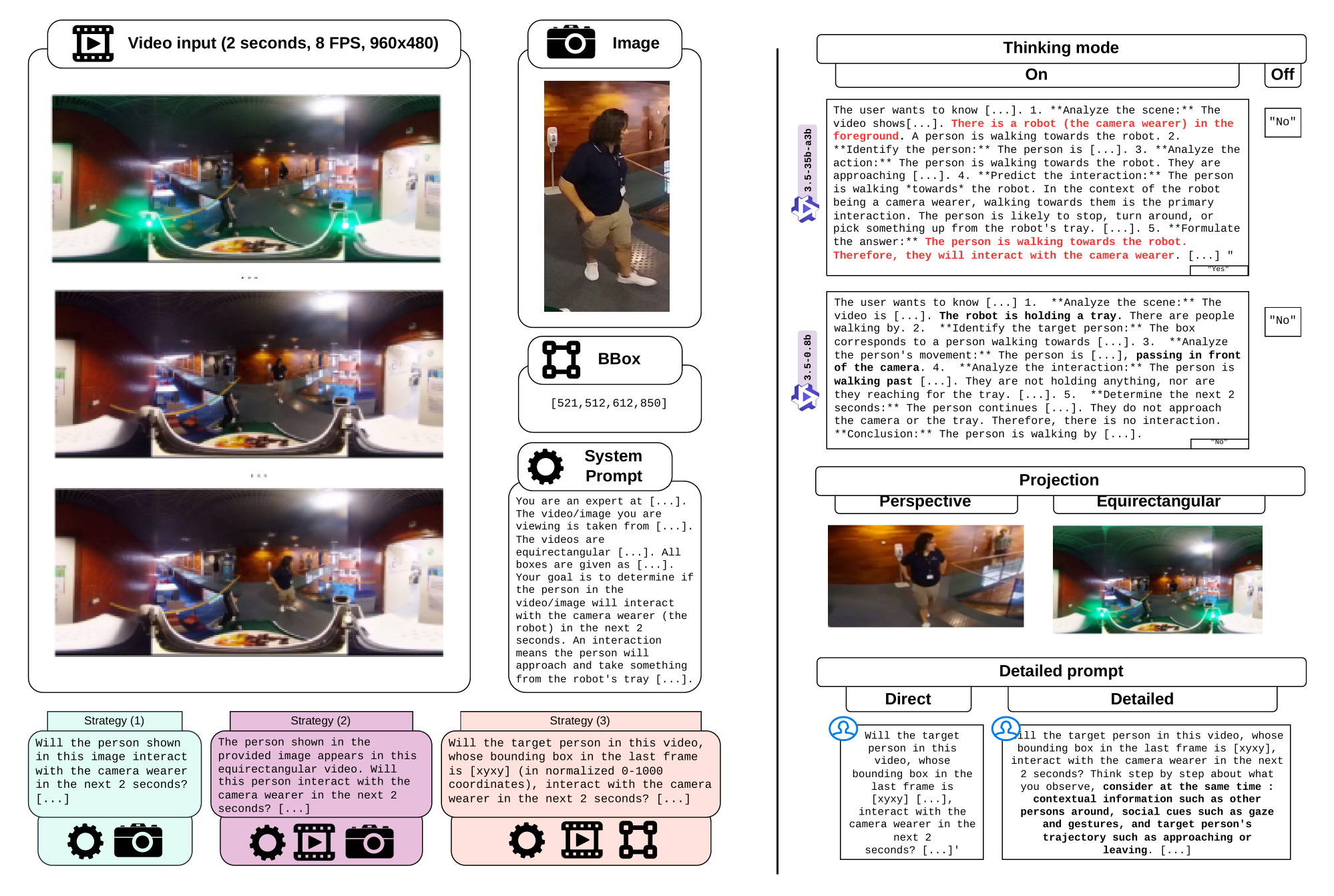}
  \caption{VLM prompting setup (left) and ablation illustration (right).
  % Prompts first reference the target with : \textit{"Will the person shown in this
  % image..."} ($\mathbf{I}_c$) \textit{"The person shown in the provided image
  % appears in this equirectangular video.
  % Will this person..."} ($\mathbf{V}+\mathbf{I}_c$) or \textit{"Will the target person in this video, whose
  % bounding box in the last frame is [x, y, x,
  % y] (in normalized 0-1000 coordinates)..."} ($\mathbf{V_{bbox}}$). Then prompts contain the question and answer format :
  % \textit{"... interact with the camera wearer in the next 2 seconds? Answer by yes or no only. Example: 'Yes.' or 'No."}
  }
  \label{fig:vlm-setup}
\end{figure}

% interact with the
%   camera wearer in the next 2
%   seconds?

We report per-track latency: $\mathbf{Pose}$ models inference time on a 32-frame sequence (excluding 2D-pose extraction); VLMs as mean API or on-device processing time per track.

\section{Preliminary Results}

Table~\ref{tab:preliminary} summarizes accuracy, precision, recall, and F1-score 
on the 100-track subset.

\textbf{$\mathbf{Pose}$ modality.}
ST-GCN achieves the best F1-score ($0.49$), ahead of SkateFormer ($0.46$) and LSTM ($0.41$).
Human annotators with the same modalities reach a mean F1 of $0.57 \pm 0.02$ (best: $0.59$), outperforming all trained pose models showing the ability of humans
to infer subtle cues even from the coarse body dynamics alone. 

\begin{figure}[t]
  \centering
  \includegraphics[width=\linewidth]{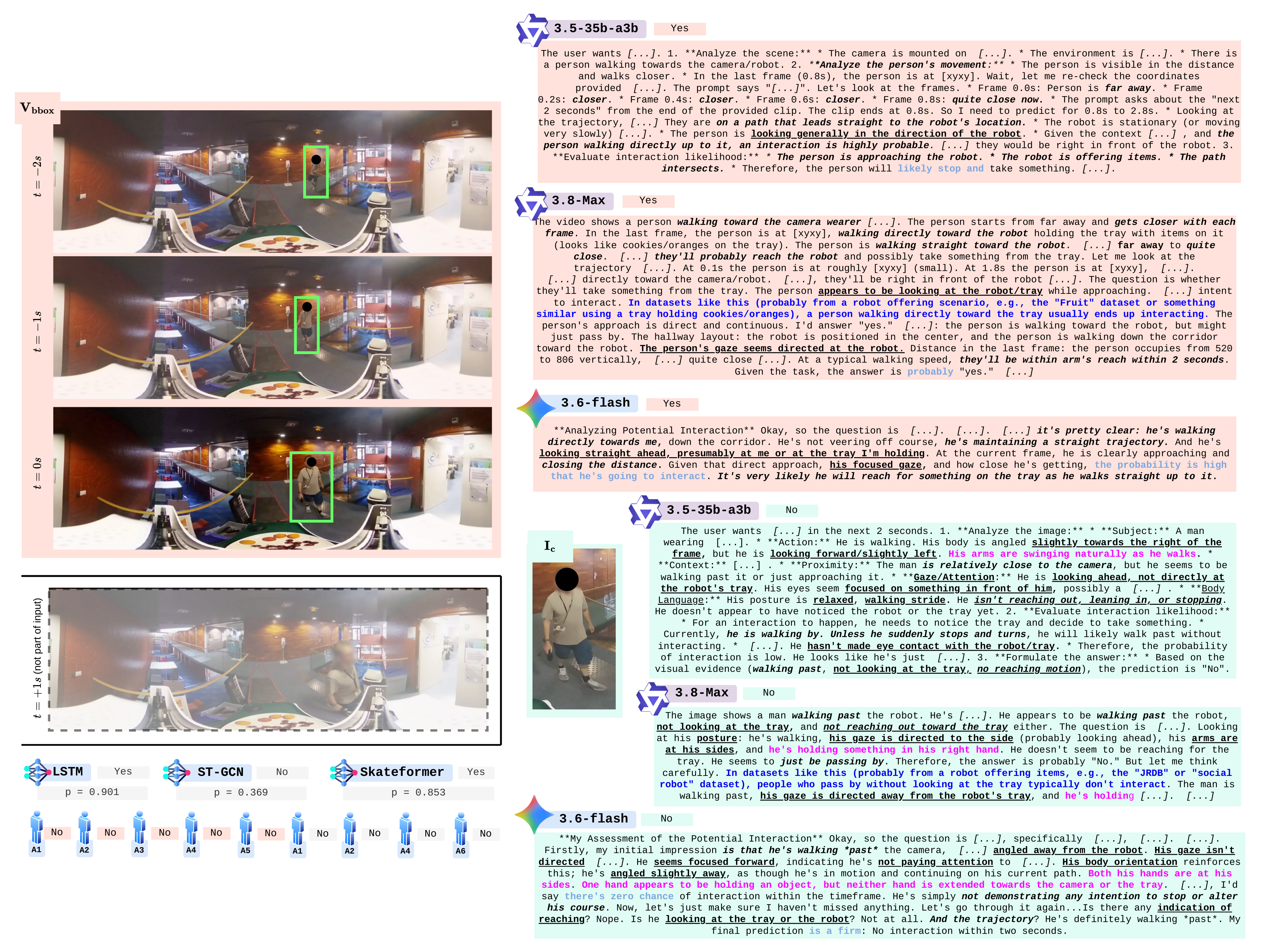}
  \caption{Qualitative examples of successful and unsuccessful predictions. Highlighted elements in the VLMs answers are 
  motion-related thinking (\textbf{\textit{bold, italic}}), visual-social related (\textbf{\underline{bold, underline}}), and other noticeable recurring elements in color: 
  certainty/uncertainty assessment (light blue), explicitly refer to related datasets (dark blue), reason about hands/arms positions (pink).}
  \label{fig:qualitative}
\end{figure}

\textbf{Visual modalities.}
$\mathbf{Pose}$ already encodes gait and approach.
Video additionally makes nonverbal cues available: gaze toward the tray, head and body orientation, occupied hands, and whether the trajectory is a pass-by or a stop.
Humans use these cues, reaching mean F1 $0.77 \pm 0.08$ under $\mathbf{V_{bbox}}$ (best: $0.90$).
VLMs given the same video often do not: the strongest zero-shot result is Qwen3.5-35B-A3B on a single crop $\mathbf{I}_c$ (F1 $0.57$), and several larger models \emph{drop} 
when temporal context is added, which is counter-intuitive for anticipation.
As illustrated in Fig.~\ref{fig:qualitative}, video traces frequently reduce the task to whether the person is approaching, yielding high-recall, 
low-precision predictions : blatant for InternVL3-1B and still visible for Qwen3.8-Max and Gemini~3.6-Flash under $\mathbf{V_{bbox}}$.
It is obvious that the video contains more information than the skeleton, and it is confirmed, and quantified, independantly of the models by the human annotators.
Then the underperformance of the VLMs highlight a important limitation of current models to reason simultaneously about temporal, spatial and social cues.

Across Table~\ref{tab:preliminary}, smaller on-device models (Qwen3.5-0.8B, InternVL3-1B) score poorly on $\mathbf{I}_c$ (F1 $0.23$--$0.35$), showing an inability to process visual social cues, 
but recover more information from dynamics with video (F1 $0.40$-$0.44$ under $\mathbf{V_{bbox}}$). 
Video does not consistently help larger models: Gemini~3.6-Flash peaks on $\mathbf{V}+\mathbf{I}_c$ (F1 $0.53$), 
whereas Qwen3.5-35B-A3B and Qwen3.8-Max perform best with $\mathbf{I}_c$.

\begin{table}[H]
  \centering
  \caption{Metrics and human baselines on 100 HUI360-Test tracks (25 pos.\ / 75 neg.): accuracy (Acc), precision (Prec), recall (Rec), and F1-Score.
  % Human rows report mean accuracy, precision, recall, and F1 across annotators; latency is mean answering time per track.
  VLM output tokens include thinking tokens; latency in seconds per track (excluding 2D-pose extraction).}
  \label{tab:preliminary}
  \scriptsize
  \setlength{\tabcolsep}{1.2pt}
  \resizebox{\columnwidth}{!}{%
  \begin{tabular}{@{}llccccccc@{}}
    \toprule
    \textbf{Method} & \textbf{Input} & \textbf{Acc} & \textbf{Prec} & \textbf{Rec} & \textbf{F1} & \textbf{Params} & \textbf{Lat.} & \textbf{Tok.} \\
    \midrule
    Humans ($n{=}4$) & $\mathbf{Pose}$ & 0.79 & 0.58 & 0.57 & 0.57 & --- & \textbf{4.3} & --- \\
    Humans ($n{=}5$) & $\mathbf{V_{bbox}}$ & \textbf{0.87} & \textbf{0.71} & \textbf{0.86} & \textbf{0.77} & --- & 4.4 & --- \\
    \midrule
    LSTM & $\mathbf{Pose}$ & \textbf{0.71} & \textbf{0.42} & 0.40 & 0.41 & \textbf{0.37M} & \textbf{0.001} & --- \\
    ST-GCN & $\mathbf{Pose}$ & 0.63 & 0.38 & 0.72 & \textbf{0.49} & 3.07M & 0.005 & --- \\
    SkateFormer & $\mathbf{Pose}$ & 0.42 & 0.30 & \textbf{1.00} & 0.46 & 1.91M & 0.017 & --- \\
    \midrule
    Qwen3.5-0.8B & $\mathbf{I}_c$ & 0.66 & 0.26 & 0.20 & 0.23 & \textbf{0.8B} & 16.1 & 611 \\
    Qwen3.5-0.8B & $\mathbf{V}+\mathbf{I}_c$ & 0.69 & 0.25 & 0.12 & 0.16 & 0.8B & 12.7 & 481 \\
    Qwen3.5-0.8B & $\mathbf{V_{bbox}}$ & 0.57 & 0.33 & 0.68 & 0.44 & 0.8B & 8.7 & 300 \\
    InternVL3-1B & $\mathbf{I}_c$ & 0.41 & 0.24 & 0.64 & 0.35 & 1B & 4.7 & 279 \\
    InternVL3-1B & $\mathbf{V}+\mathbf{I}_c$ & 0.29 & 0.26 & \textbf{1.00} & 0.41 & 1B & 3.2 & 82 \\
    InternVL3-1B & $\mathbf{V_{bbox}}$ & 0.25 & 0.25 & \textbf{1.00} & 0.40 & 1B & \textbf{2.7} & 80 \\
    Qwen3.5-35B-A3B & $\mathbf{I}_c$ & \textbf{0.79} & \textbf{0.58} & 0.56 & \textbf{0.57} & 35B & 3.4 & 521 \\
    Qwen3.5-35B-A3B & $\mathbf{V}+\mathbf{I}_c$ & 0.71 & 0.44 & 0.64 & 0.52 & 35B & 5.2 & 649 \\
    Qwen3.5-35B-A3B & $\mathbf{V_{bbox}}$ & 0.60 & 0.35 & 0.72 & 0.47 & 35B & 6.2 & 819 \\
    Qwen3.8-Max & $\mathbf{I}_c$ & 0.70 & 0.42 & 0.56 & 0.48 & 2.4T & 11.7 & 535 \\
    Qwen3.8-Max & $\mathbf{V}+\mathbf{I}_c$ & 0.63 & 0.35 & 0.56 & 0.43 & 2.4T & 17.0 & 699 \\
    Qwen3.8-Max & $\mathbf{V_{bbox}}$ & 0.54 & 0.33 & 0.84 & 0.48 & 2.4T & 16.3 & 684 \\
    Gemini~3.5-Flash-L & $\mathbf{I}_c$ & 0.73 & 0.46 & 0.52 & 0.49 & --- & \textbf{2.5} & 135 \\
    Gemini~3.5-Flash-L & $\mathbf{V}+\mathbf{I}_c$ & 0.65 & 0.37 & 0.56 & 0.44 & --- & 3.6 & 144 \\
    Gemini~3.5-Flash-L & $\mathbf{V_{bbox}}$ & 0.56 & 0.31 & 0.60 & 0.41 & --- & 3.7 & 172 \\
    Gemini~3.6-Flash & $\mathbf{I}_c$ & 0.69 & 0.42 & 0.64 & 0.51 & --- & 4.7 & 267 \\
    Gemini~3.6-Flash & $\mathbf{V}+\mathbf{I}_c$ & 0.68 & 0.42 & 0.72 & 0.53 & --- & 6.5 & 367 \\
    Gemini~3.6-Flash & $\mathbf{V_{bbox}}$ & 0.53 & 0.31 & 0.72 & 0.43 & --- & 6.1 & 363 \\
    \bottomrule
  \end{tabular}}
\end{table}

\FloatBarrier

\begin{figure}[H]
  \centering
  \includegraphics[width=0.9\linewidth]{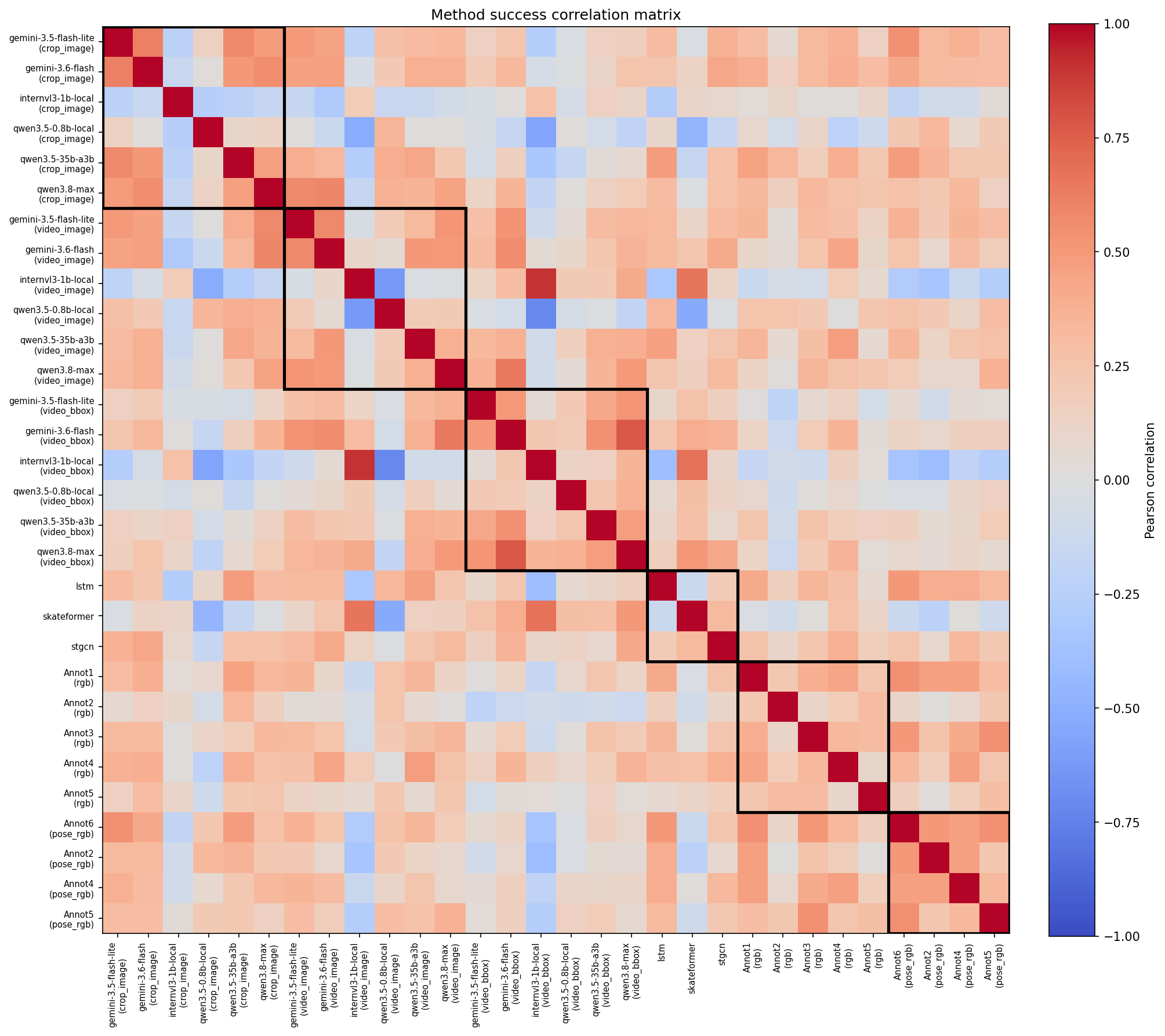}
  \caption{Pearson correlation ($r$) between the predictions of all methods and human annotators. Diagonal blocks highlight the correlations between the same family of methods
  under the same modalities.}
  \label{fig:correlation}
\end{figure}

The Pearson correlation ($r$) between all methods (Fig.~\ref{fig:correlation}) reflects the difference between larger and smaller models when analyzing
the image and the subtle cues: large models show a relatively high positive $r$ in the range $0.38-0.62$ (same family) or $0.32-0.78$ (across families).
Under the $\mathbf{V_{bbox}}$ modality, as the focus is given on higher level analysis of the 
approach motion, the correlation remains positive ($r\in [0.05, 0.21]$) between larger and smaller models (InternVL3-1B and Qwen3.5-0.8B).
It drops for Qwen3.5-0.8B and even turns negative for InternVL3-1B under $\mathbf{I}_c$ and $\mathbf{V}+\mathbf{I}_c$ ($r\in[-0.24, 0.11]$).
This trend of motion cues leading to more agreement than social visual cues also holds for humans, whose average $r$ is $0.42$ under the skeletal modality
and $0.26$ under the visual one, highlighting a higher individual bias when human annotators have access to the visual appearance of the target person.

\FloatBarrier

\textbf{Ablations.}
Table~\ref{tab:ablations} reports targeted comparisons on Qwen3.5-0.8B and Qwen3.5-35B-A3B under $\mathbf{V_{bbox}}$.
Disabling thinking mode confirms its necessity, as it yields near-zero F1 because models default to almost always answering ``No'' with very short outputs.
Increasing video sampling from 2 to 8~FPS improves F1 monotonically for both models.
Equirectangular input consistently outperforms a pinhole-camera style reprojection ($90 \times 60$ degrees centered on the target person)
of the same clip suggesting that the whole visual context, and in particular having the target interaction zone visible is necessary.
To encourage the model to reason over both dynamics and social cues we elaborated a longer prompt (\texttt{detailed}) but under thinking mode 
it has a modest effect, and the short format is slightly preferred.

We saw that the thinking process for Qwen3.8-Max was occasionally interrupted so we 
vary the reasoning token budget under $\mathbf{V_{bbox}}$.
Increasing the budget from 1024 to 8192 tokens raises F1 from $0.48$ to $0.51$ and recall from $0.84$ to $0.96$.

\begin{table}[t]
  \centering
  \caption{VLM ablations on $\mathbf{V_{bbox}}$ (100 tracks). Settings marked with $^*$ are defaults.
  For Qwen3.8-Max (bottom block), latency and token columns report mean $\pm$ std.}
  \label{tab:ablations}
  \footnotesize
  \setlength{\tabcolsep}{2pt}
  \resizebox{\columnwidth}{!}{%
  \begin{tabular}{@{}llcccccccc@{}}
    \toprule
    \multirow{2}{*}{\textbf{Ablation}} & \multirow{2}{*}{\textbf{Setting}}
    & \multicolumn{4}{c}{\textbf{Qwen3.5-0.8B}}
    & \multicolumn{4}{c}{\textbf{Qwen3.5-35B-A3B}} \\
    \cmidrule(lr){3-6}\cmidrule(lr){7-10}
    & & \textbf{F1} & \textbf{Rec} & \textbf{Lat.} & \textbf{Tok.}
    & \textbf{F1} & \textbf{Rec} & \textbf{Lat.} & \textbf{Tok.} \\
    \midrule
    \multirow{2}{*}{Thinking} & without & 0.08 & 0.04 & \textbf{0.2} & \textbf{3} & 0.42 & 0.36 & \textbf{1.4} & \textbf{3} \\
    & with$^*$ & \textbf{0.44} & \textbf{0.68} & 8.7 & 300 & \textbf{0.47} & \textbf{0.72} & 6.2 & 819 \\
    \midrule
    \multirow{3}{*}{FPS} & 2 & 0.30 & 0.40 & 13.1 & 345 & 0.37 & 0.44 & \textbf{4.7} & \textbf{754} \\
    & 4 & 0.30 & 0.40 & 9.0 & 345 & 0.39 & 0.52 & 5.0 & 784 \\
    & 8$^*$ & \textbf{0.44} & \textbf{0.68} & \textbf{8.7} & \textbf{300} & \textbf{0.47} & \textbf{0.72} & 6.2 & 819 \\
    \midrule
    \multirow{2}{*}{Prompt} & detailed & 0.36 & 0.56 & 14.4 & 559 & \textbf{0.47} & \textbf{0.76} & 6.7 & 960 \\
    & direct$^*$ & \textbf{0.44} & \textbf{0.68} & \textbf{8.7} & \textbf{300} & \textbf{0.47} & 0.72 & \textbf{6.2} & \textbf{819} \\
    \midrule
    \multirow{2}{*}{Projection} & pinhole & 0.16 & 0.12 & 11.2 & 383 & 0.38 & 0.44 & 5.0 & 610 \\
    & equirectangular$^*$ & \textbf{0.44} & \textbf{0.68} & \textbf{8.7} & \textbf{300} & \textbf{0.47} & \textbf{0.72} & \textbf{6.2} & \textbf{819} \\
    \midrule
    & & \multicolumn{8}{c}{\textbf{Qwen3.8-Max}} \\
    \cmidrule(lr){3-10}
    & & \multicolumn{2}{c}{\textbf{F1}} & \multicolumn{2}{c}{\textbf{Rec}}
    & \multicolumn{2}{c}{\textbf{Lat.}} & \multicolumn{2}{c}{\textbf{Tok.}} \\
    \midrule
    \multirow{3}{*}{Thinking budget} & 1024$^*$ & \multicolumn{2}{c}{0.48} & \multicolumn{2}{c}{0.84}
    & \multicolumn{2}{c}{\textbf{16.3} $\pm$ 6.4} & \multicolumn{2}{c}{\textbf{684} $\pm$ 315} \\
    & 2048 & \multicolumn{2}{c}{0.49} & \multicolumn{2}{c}{0.88}
    & \multicolumn{2}{c}{23.3 $\pm$ 12.0} & \multicolumn{2}{c}{908 $\pm$ 636} \\
    & 8192 & \multicolumn{2}{c}{\textbf{0.51}} & \multicolumn{2}{c}{\textbf{0.96}}
    & \multicolumn{2}{c}{24.5 $\pm$ 22.3} & \multicolumn{2}{c}{1101 $\pm$ 1240} \\
    \bottomrule
  \end{tabular}}
\end{table}

\section{Conclusion and Discussion}

Humans are able to anticipate interaction from egocentric cues with little effort although not perfectly: they 
outperform pose models on $\mathbf{Pose}$ alone and gain substantially 
from access to the full video, whereas VLMs struggle to reason over social dynamics 
and often reduce the task to approach detection or focus on irrelevant details despite thinking mode. 
We observe no clear correlation between model size and performance, and trained pose models such as ST-GCN remain an acceptable lightweight 
alternative when compute is limited. These results are preliminary; future work should compare VLMs to lighter image- 
or video-based models trained on HUI360, probe whether failures occur at social perception or at the anticipation decision, 
and conduct a deeper user study to separate rational inference from intuitive judgment in human annotators.

\footnotesize
\section*{Acknowledgment}
This work was partially supported by the France
2030 program through the PEPR O2R projects AS3 (ANR-22-EXOD-007)
and the ENACT AI Cluster (ANR-23-IACL-0004).
This work was performed using HPC resources from GENCI-IDRIS (Grant 2026-AD011016950).

\bibliographystyle{IEEEtran}
\bibliography{references}

\end{document}